\documentclass[conference]{IEEEtran}
    \IEEEoverridecommandlockouts
    \usepackage{cite}
    \usepackage{amsmath,amssymb,amsfonts}
    \usepackage{algorithmic}
    \usepackage{graphicx}
    \usepackage{textcomp}
    \usepackage{xcolor}
    \usepackage{comment}
    \usepackage{url}
    \usepackage{booktabs}
    \usepackage{multirow}
    \def\BibTeX{{\rm B\kern-.05em{\sc i\kern-.025em b}\kern-.08em
        T\kern-.1667em\lower.7ex\hbox{E}\kern-.125emX}}
    
    \usepackage[hidelinks]{hyperref}

\begin{document}
    
    \title{A Cross-Domain Evaluation of Prompting Techniques for Large Language Models}
    
    \author{\IEEEauthorblockN{Kevin Krazius, Tim Schlippe, Paul Libbrecht}
    \IEEEauthorblockA{
    \textit{IU International University of Applied Sciences}\\
    Germany \\
    tim.schlippe@iu.org}
    }

\maketitle
    
\begin{abstract}
Prompt engineering has become a central mechanism for guiding the behavior of Large Language Models (LLMs) across a wide range of reasoning and language understanding tasks \cite{schulhoff2025promptreport,liu2023pretrain}. However, existing studies frequently evaluate prompting techniques using different LLMs, datasets, or task settings, making direct comparisons across methods and domains difficult \cite{chang2024survey}. Consequently, in this paper we present a controlled cross-domain evaluation of Zero-Shot prompting, Few-Shot prompting, Retrieval-Augmented Generation (RAG), and automatic prompt optimization based on a PE2-inspired optimization pipeline \cite{brown2020languagemodels,lewis2020rag,ye2024promptengineer}. All experiments were conducted with the same LLM and evaluated on six benchmark datasets covering arithmetic reasoning, logical reasoning, commonsense reasoning, and factual verification. 
The results show that no prompting strategy consistently outperforms the others across all domains, and pairwise statistical comparisons of the highest-performing configurations revealed no significant differences. RAG achieves the highest overall average accuracy of 78.33\% when 300 examples are available for retrieval, while Few-Shot Prompting achieves comparatively high accuracies across several structured reasoning tasks. Zero-Shot Prompting remains competitive on tasks with self-contained inputs, whereas the PE2-inspired optimization approach produces stable but not systematically superior results compared to the other evaluated prompting techniques. Furthermore, increasing the number of task-specific examples does not consistently improve performance across prompting techniques, indicating strong domain and task dependency. 
These findings suggest that prompting techniques should be selected adaptively based on task characteristics rather than treated as universally applicable optimization strategies.
\end{abstract}
    
    
\section{Introduction}
    
Large Language Models (LLMs) have fundamentally transformed Natural Language Processing (NLP) by enabling strong performance across a broad range of language understanding and reasoning tasks without task-specific parameter updates or supervised fine-tuning \cite{brown2020languagemodels,openai2023gpt4}. Instead of modifying model parameters for individual downstream tasks, many modern approaches rely on prompting techniques that guide LLM behavior through natural language instructions and contextual examples \cite{liu2023pretrain,schulhoff2025promptreport}.
    
Among the most widely used prompting techniques are Zero-Shot Prompting, Few-Shot Prompting, Retrieval-Augmented Generation (RAG), and automatic prompt optimization approaches \cite{wei2022chain,fengEtAlRAGs,ye2024promptengineer}. Zero-Shot Prompting relies exclusively on task instructions without providing demonstration examples, whereas Few-Shot Prompting incorporates example-based demonstrations directly into the prompt context \cite{brown2020languagemodels}. RAG extends the prompting process by retrieving external knowledge or task-relevant context before response generation, thereby improving access to information beyond the LLM's parametric knowledge \cite{lewis2020rag,Kivroglou2025InvestigatingRetrieval22wm,Schlippe2026RAGStrategies6i2k}.  More recently, automatic prompt optimization approaches have attempted to improve prompt quality algorithmically instead of relying solely on manual prompt engineering \cite{ye2024promptengineer}.
    
Although numerous studies report significant performance improvements through advanced prompting strategies, the comparability of existing results remains limited: Many evaluations use different LLMs, benchmark datasets, evaluation metrics, and experimental configurations, making it difficult to isolate the actual impact of individual prompting techniques \cite{chang2024survey}. Furthermore, several studies focus primarily on a single reasoning domain, such as mathematical reasoning or factual question answering, rather than evaluating prompting techniques consistently across multiple reasoning tasks.
    
Consequently, our paper addresses these limitations through a controlled cross-domain evaluation of prompting techniques using a unified experimental setup. GPT-4o-mini was used as a representative LLM across all experiments to isolate differences between prompting techniques from differences between LLMs. Identical evaluation procedures were applied across six benchmark datasets covering arithmetic reasoning, logical reasoning, commonsense reasoning, and factual verification tasks. The evaluated prompting techniques include Zero-Shot Prompting, Few-Shot Prompting, Retrieval-Augmented Generation, and a PE2-inspired automatic prompt optimization pipeline. These techniques were selected as they represent four fundamentally different prompting paradigms: instruction-only prompting, example-based prompting, retrieval-augmented prompting, and automatic prompt optimization.
    
The primary objective of our paper is to analyze how different prompting techniques influence LLM performance across reasoning domains and whether more complex prompting approaches consistently outperform simpler prompting strategies. In addition, our paper investigates the extent to which automatically optimized prompts generalize across benchmark datasets compared to manually designed prompts.
    
Our main contributions are summarized as follows:
    
\begin{itemize}
    \item A controlled comparative evaluation of multiple prompting techniques using the same LLM and a unified experimental setup.
        
    \item A cross-domain benchmark analysis covering arithmetic reasoning, logical reasoning, commonsense reasoning, and factual verification tasks.
        
    \item An empirical evaluation of a PE2-inspired \cite{ye2024promptengineer} automatic prompt optimization pipeline.
        
    \item An analysis of the relationship between prompt complexity, example-based context integration, and reasoning performance.
\end{itemize}
    
To support reproducibility and facilitate future research, all experimental results generated in this study have been made publicly available\footnote{\url{https://github.com/dieKrake/prompting-techniques}}.
    
The remainder of this paper is structured as follows: Section~\ref{sec:relatedwork} reviews related work on LLMs, prompting techniques, and prompt optimization. Section~\ref{sec:methodology} describes the experimental setup, benchmark datasets, and evaluation methodology. Section~\ref{sec:results} presents the experimental results and cross-domain analysis. Finally, Section~\ref{sec:conclusion} summarizes the findings, discusses limitations, and outlines directions for future research.
    
\section{Related Work}
\label{sec:relatedwork}

This section reviews prior work on prompting techniques, retrieval-augmented generation, and automatic prompt optimization and subsequently summarizes the research gap addressed in our study.
    
\subsection{Prompting Techniques}
    
Prompting has become a central approach for adapting LLM behavior without updating LLM parameters \cite{brown2020languagemodels,liu2023pretrain,schulhoff2025promptreport}. In contrast to supervised fine-tuning, prompting relies on task descriptions, contextual information, and examples provided at inference time \cite{brown2020languagemodels}. Zero-Shot Prompting represents the simplest prompting setting, where the LLM receives only a task instruction and the input instance \cite{brown2020languagemodels}. Few-Shot Prompting extends this setup by adding demonstration examples to the prompt context, enabling the LLM to infer the desired task format from examples \cite{brown2020languagemodels}. Prior work has shown that the choice, order, and relevance of examples can substantially affect Few-Shot Prompting performance \cite{liu2022knn}.
    
\subsection{Retrieval-Augmented Generation}
    
RAG combines LLM prompting with information retrieval by adding retrieved context to the prompt before response generation \cite{lewis2020rag}. This approach is particularly relevant for knowledge-intensive tasks as it allows the LLM to condition its response on external information instead of relying only on parametric knowledge \cite{lewis2020rag}. Retrieval-based LLM approaches have shown that retrieved passages or examples can improve factual grounding and task performance, depending on the quality and relevance of the retrieved context \cite{lewis2020rag,borgeaud2022retro}. While a broad diversity of RAG approaches exists~\cite{fengEtAlRAGs}, this study evaluates an example-based RAG configuration in which semantically relevant examples are dynamically retrieved from a task-specific example pool and added to the prompt context.
    
\subsection{Automatic Prompt Optimization}
    
Automatic prompt optimization aims to reduce manual prompt engineering by generating, modifying, and evaluating prompts algorithmically \cite{zhou2022humanlevel,pryzant2023automatic,ye2024promptengineer}. The Automatic Prompt Engineer (APE) approach showed that LLMs can generate and rank candidate instructions for downstream tasks \cite{zhou2022humanlevel}. The ProTeGi approach introduced an optimization procedure that uses textual gradients and beam search to iteratively improve prompts based on observed errors \cite{pryzant2023automatic}. PE2 further investigates the use of prompt engineering methods to construct improved prompt engineers \cite{ye2024promptengineer}. Since the original PE2 implementation is not publicly available, our paper evaluates a PE2-inspired optimization pipeline that follows the same general idea of iterative error-based prompt improvement.
    
\subsection{Research Gaps}
    
Despite extensive work on individual prompting techniques, direct comparisons between Zero-Shot Prompting, Few-Shot Prompting, RAG, and automatic prompt optimization under identical experimental conditions remain limited \cite{chang2024survey,schulhoff2025promptreport}. In particular, many studies focus on individual reasoning domains or evaluate prompting techniques using different experimental setups, making it difficult to determine whether observed performance differences result from the prompting technique itself or from dataset-specific and setup-specific factors \cite{chang2024survey}. To address this gap, our paper evaluates multiple prompting techniques using the same LLM, the same benchmark datasets, and the same evaluation metric across several reasoning domains.

\section{Methodology}
\label{sec:methodology}
    
\begin{figure}[h!]
\centering
\includegraphics[width=0.5\textwidth]{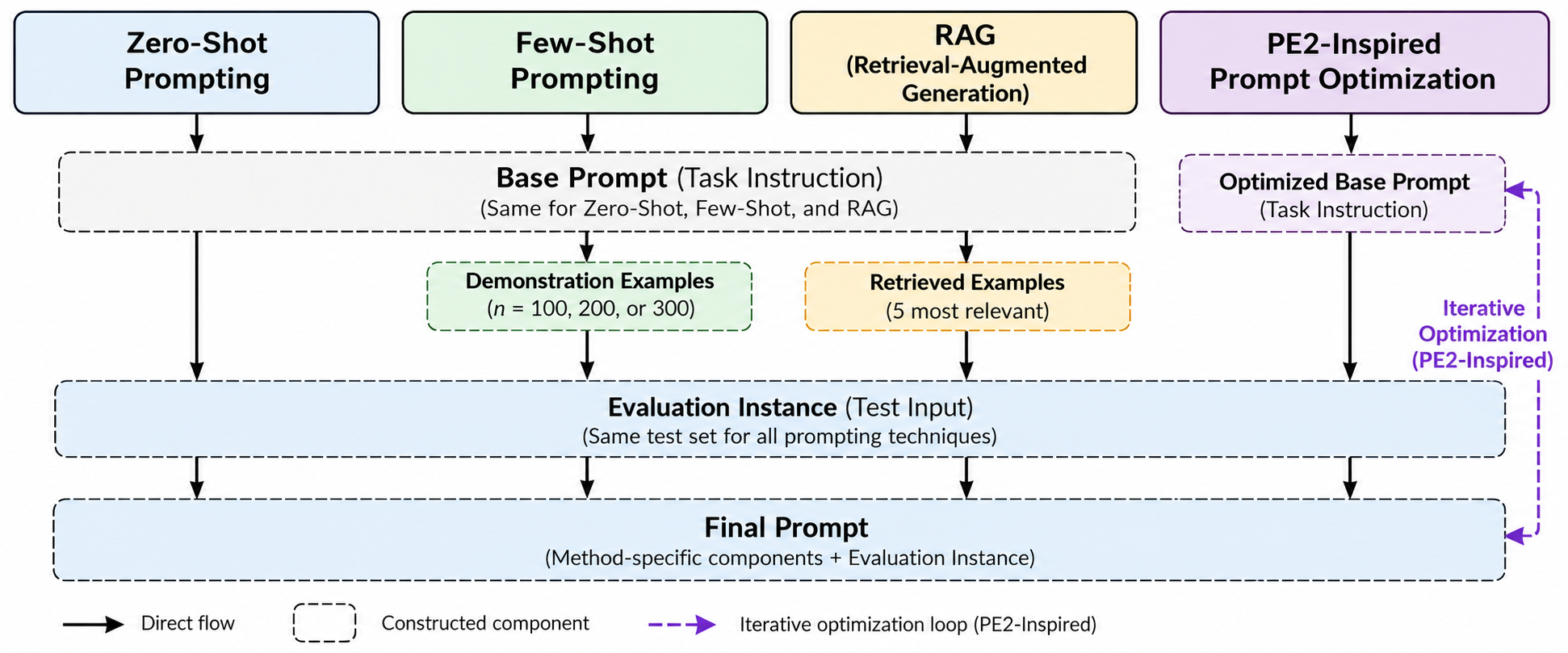}
\caption{Overview of the evaluated prompting techniques and prompt construction workflow.}
\label{fig:overview_prompting}
\end{figure}

This section describes the experimental setup, benchmark datasets, prompting techniques, and evaluation procedure used in this paper. Figure~\ref{fig:overview_prompting} summarizes how the evaluated prompting techniques construct the final prompt used during inference. Zero-Shot Prompting relies only on the base prompt, Few-Shot Prompting extends the base prompt with fixed demonstration examples, and RAG dynamically retrieves semantically relevant examples from a pool of task-specific examples. PE2 differs from the other approaches by iteratively optimizing the base prompt before evaluation. For all prompting techniques, the resulting method-specific components are combined with the same evaluation instance.
    
    \subsection{LLM Configuration}
    
    We conducted all experiments using GPT-4o-mini through the OpenAI Chat Completions API \cite{openai2024gpt4o}. The LLM was selected as it provides strong reasoning capabilities while maintaining efficient inference performance across large benchmark evaluations.
    
    To ensure comparability across prompting techniques, identical inference settings were used for all experiments. Responses were generated with a temperature of 0.0, and each benchmark instance was processed through isolated API requests to avoid unintended context leakage between evaluation samples.
    
    The experimental pipeline was implemented in Python using the OpenAI API and the HuggingFace Datasets library \cite{lhoest2021datasets}.
    
    \subsection{Benchmark Datasets}
    
    The prompting techniques were evaluated on six benchmark datasets covering multiple reasoning domains:
    
    \begin{itemize}
    \item \textbf{Elementary Math QA} for arithmetic and mathematical reasoning tasks \cite{srivastava2023beyond}
    
    \item \textbf{Logical Fallacy Detection} for logical reasoning and fallacy classification tasks \cite{srivastava2023beyond}
    
    \item \textbf{Navigate} for spatial and abstract logical reasoning tasks \cite{srivastava2023beyond}
        
        \item \textbf{BoolQ} for boolean question answering and reading comprehension \cite{clark2019boolq}
        
        \item \textbf{Social IQA} for social commonsense reasoning and understanding of human interactions \cite{sap2019socialiqa}
        
    \item \textbf{Fact Checker} for factual verification and fact-based reasoning tasks \cite{srivastava2023beyond}
    \end{itemize}
    
For each dataset, 200 test instances were used consistently across all prompting techniques to enable controlled within-dataset comparisons while keeping the number of API evaluations manageable across the full experimental setup. These test instances were kept separate from the additional task-specific examples used by the prompting techniques. Separate sets of 100, 200, and 300 examples were used as demonstration examples for Few-Shot Prompting, as examples available for retrieval for RAG, and during prompt optimization for PE2. These three sizes were selected to enable a controlled comparison of how the amount of available task-specific examples affects the evaluated prompting techniques while keeping the experimental scope manageable.
    
The selected datasets cover arithmetic reasoning, logical reasoning, commonsense reasoning, reading comprehension, and factual verification to evaluate prompting behavior in diverse reasoning domains.
    
The datasets derived from BIG-bench were based on the publicly available task implementations provided in the BIG-bench benchmark suite \cite{srivastava2023beyond}.

\subsection{Zero-Shot Prompting}
    
As illustrated in Figure~\ref{fig:overview_prompting}, Zero-Shot Prompting used only a task-specific base prompt together with the evaluation instance. No demonstration examples were added to the prompt context.
    
The prompts were designed to constrain the expected output format and reduce ambiguous LLM outputs. Depending on the dataset, the prompts instructed the LLM to return binary decisions, multiple-choice answers, or concise factual outputs without additional explanations.
    
For example, the Navigate dataset used a restrictive binary classification prompt requiring the LLM to answer exclusively with ``True'' or ``False'', while Elementary Math QA used a multiple-choice instruction format requiring the LLM to output only the correct answer option.
    
\subsection{Few-Shot Prompting}
    
Few-Shot Prompting extended the base prompt with examples sampled from the corresponding dataset. The examples followed the same input-output structure as the evaluation task and were inserted before the evaluation instance within the prompt context.
    
Unlike Zero-Shot Prompting, the LLM therefore received demonstrations of expected task behavior and output structure. The examples were selected from the available dataset subsets and remained fixed during evaluation.
    
\begin{figure}[h!]
\centering
\includegraphics[width=\columnwidth]{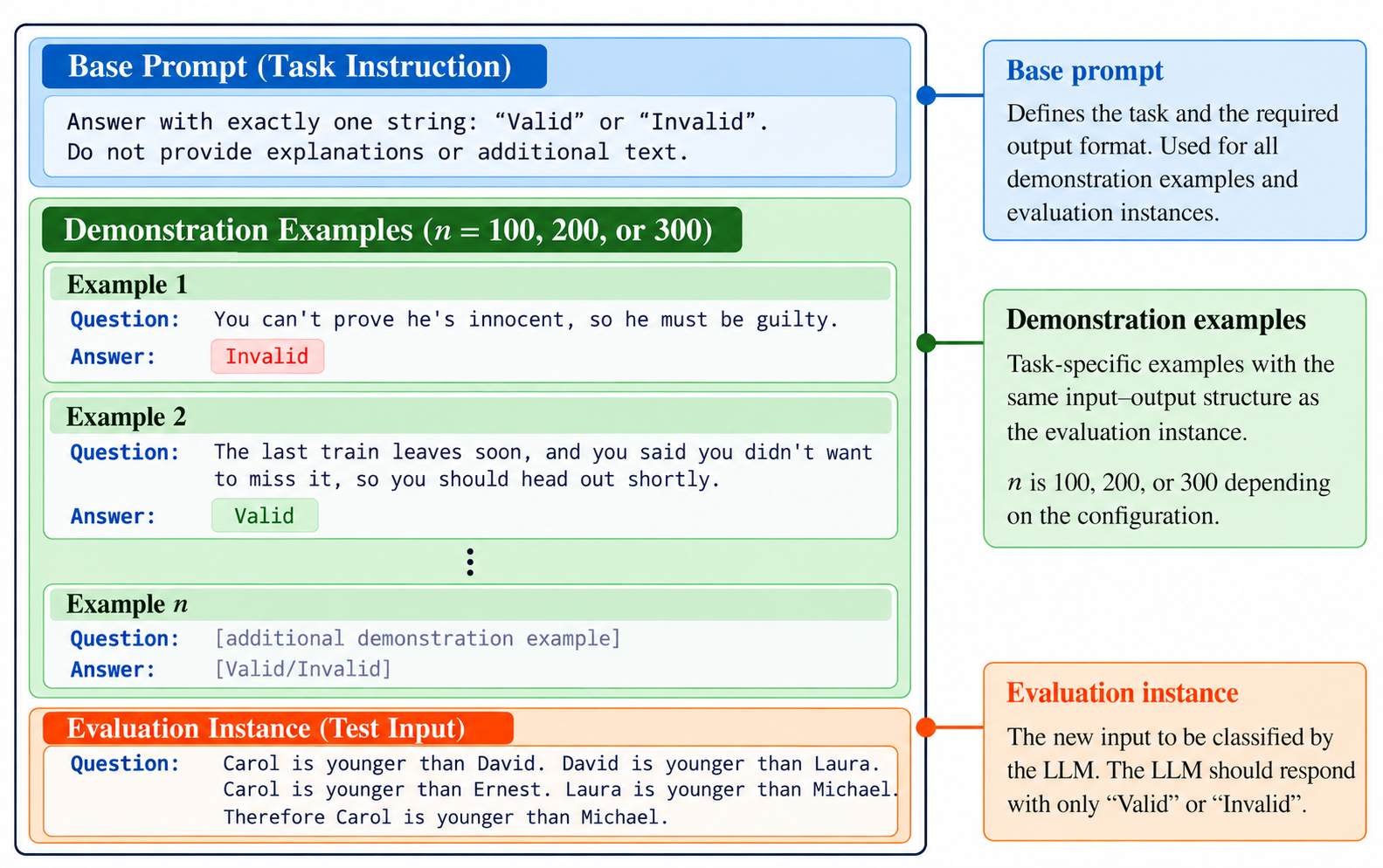}
\caption{Few-Shot prompt structure for Logical Fallacy Detection.}
\label{fig:fewshot_prompt}
\end{figure}

The Few-Shot prompt consisted of the base prompt, fixed demonstration examples, and the evaluation instance. Figure~\ref{fig:fewshot_prompt} illustrates this structure using actual examples from the Logical Fallacy Detection experiments.
    
    In the complete Few-Shot prompts, $n$ was 100, 200, or 300 depending on the configuration; Figure~\ref{fig:fewshot_prompt} shows only two demonstration examples for readability. The complete prompts and experimental results are available in our \hyperref[repository]{repository}.

\subsection{Retrieval-Augmented Generation}
    
For RAG, semantically relevant examples were dynamically retrieved from the same task-specific example pools used to construct the Few-Shot configurations. Following related work using five retrieved in-context examples \cite{tejaswi2025rare}, we fixed the number of retrieved examples at $k=5$ across all RAG configurations.

The resulting prompt followed the same general structure illustrated in Figure~\ref{fig:fewshot_prompt}. Unlike Few-Shot Prompting, RAG selected the examples separately for each evaluation instance.
    
The retrieval pipeline used Azure AI Search with embedding-based similarity matching. For each evaluation instance, the five most similar examples were retrieved from the corresponding task-specific example pool and added to the prompt context. The size of this pool varied between 100, 200, and 300 examples across the evaluated RAG configurations.
    
Within the experimental setup of this paper, RAG therefore functioned as retrieval-based example augmentation rather than as retrieval from an external knowledge corpus. This design maintained the controlled experimental setting by using task-specific examples from the same benchmark datasets as the other example-based prompting techniques.

\subsection{Automatic Prompt Optimization}
    
Automatic prompt optimization was implemented using an iterative approach inspired by PE2 \cite{ye2024promptengineer}. As the original PE2 implementation was not publicly available, we implemented a PE2-inspired optimization procedure that iteratively used incorrectly answered examples to generate and select revised prompt variants.
    
\begin{figure}[t]
    \centering
    \includegraphics[width=0.5\textwidth]{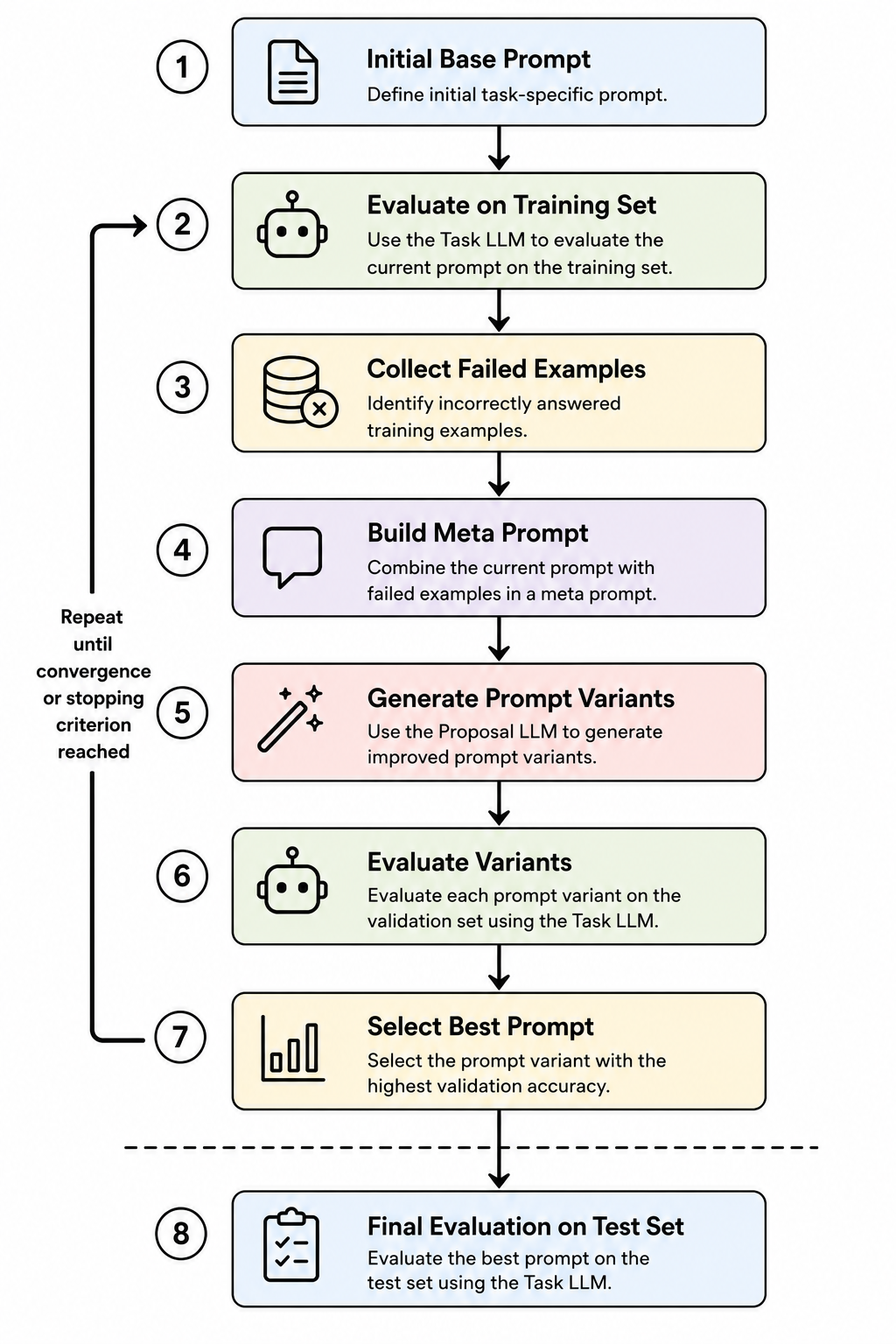}
    \caption{PE2-inspired iterative prompt optimization pipeline.}
    \label{fig:pe2_pipeline}
    \end{figure}

    Figure~\ref{fig:pe2_pipeline} illustrates the optimization workflow. The 100, 200, and 300 examples used for optimization were equally divided into training and validation sets. Training examples were used to identify failures for prompt revision, while the validation set was used to select prompt variants. The separate test set was used only for the final evaluation of the selected prompt.
    
    The optimization process consisted of the following steps:
    
    \begin{enumerate}
        \item Define an initial task-specific base prompt.
        \item Evaluate the current prompt on the training set.
        \item Collect incorrectly answered examples.
        \item Construct a meta-prompt containing the current prompt and failed examples.
        \item Generate revised prompt variants using the proposal LLM.
        \item Evaluate the generated variants on a validation set.
        \item Select the highest-performing prompt variant.
        \item Evaluate the selected prompt on the test set.
    \end{enumerate}
    
The meta-prompt contained the current prompt and incorrectly answered training examples and instructed the proposal LLM to analyze the observed failures and generate revised prompts while preserving the task and output format. Both the task and proposal roles used GPT-4o-mini. The optimization was limited to three iterations and used a beam-like selection strategy based on validation accuracy. The complete meta-prompt and PE2-inspired implementation are available in our \hyperref[repository]{repository}.
    
Unlike Few-Shot Prompting and RAG, PE2 therefore attempted to improve the prompt instructions themselves rather than modifying the contextual examples provided during inference.

\begin{table*}[h!]
    \centering
    \footnotesize
    \caption{Accuracy (\%) across prompting techniques and configurations. The best result per dataset is shown in bold.}
    \label{tab:results_combined}
    \resizebox{\textwidth}{!}{
    \begin{tabular}{lrrrrrrrrrr}
    \toprule
    & Zero-Shot & \multicolumn{3}{c}{Few-Shot} & \multicolumn{3}{c}{RAG} & \multicolumn{3}{c}{PE2} \\
    \cmidrule(lr){3-5}
    \cmidrule(lr){6-8}
    \cmidrule(lr){9-11}
    Dataset & -- & 100 & 200 & 300 & 100 & 200 & 300 & 100 & 200 & 300 \\
    \midrule
    Elementary Math QA 
    & 54.5 
    & 62.5 
    & 60.5 
    & 58.5 
    & 60.5 
    & 65.5 
    & \textbf{66.0} 
    & 51.5 
    & 56.0 
    & 53.0 \\
    
    Logical Fallacy Detection 
    & 78.0 
    & 81.5 
    & 85.0 
    & 80.0 
    & 79.0 
    & 80.0 
    & \textbf{85.5} 
    & 77.5 
    & 75.0 
    & 77.5 \\
    
    Navigate 
    & 66.5 
    & 69.5 
    & 69.5 
    & \textbf{71.5} 
    & 64.0 
    & 68.0 
    & 66.0 
    & 69.0 
    & 66.5 
    & 66.0 \\
    
    BoolQ 
    & \textbf{90.0} 
    & 88.5 
    & 88.5 
    & 88.0 
    & 89.5 
    & 89.0 
    & 87.5 
    & 87.5 
    & 88.5 
    & 89.5 \\
    
    Social IQA 
    & 77.5 
    & 78.0 
    & 77.0 
    & 78.0 
    & 77.0 
    & 74.5 
    & 78.5 
    & \textbf{79.0} 
    & 78.0 
    & 78.5 \\
    
    Fact Checker 
    & 80.5 
    & 81.5 
    & \textbf{86.5} 
    & 85.5 
    & 84.5 
    & \textbf{86.5} 
    & \textbf{86.5} 
    & 80.0 
    & 81.5 
    & 79.5 \\
    
    \midrule
    Average 
    & 74.50 
    & 76.92 
    & 77.83 
    & 76.92 
    & 75.75 
    & 77.25 
    & \textbf{78.33} 
    & 74.08 
    & 74.25 
    & 74.00 \\
    \bottomrule
    \end{tabular}}
\end{table*}

\subsection{Evaluation Procedure}
    
For each benchmark instance, the generated response was compared against the corresponding ground-truth label. Performance was measured using accuracy, defined as the proportion of correctly answered instances within the evaluation set. Accuracy was used as a common metric as each benchmark instance had a single ground-truth answer, enabling consistent performance comparisons across the different task formats.
    
The experiments focused on controlled comparisons between prompting techniques across reasoning domains rather than on estimating state-of-the-art benchmark performance. The same test instances were therefore used for all prompting techniques within each dataset.
    
To assess whether differences between the highest-performing configurations were statistically significant, we used two-sided exact McNemar tests for paired binary outcomes \cite{mcnemar1947,westfall2010multiple}. Outcomes were paired within each test instance, while the test instances were treated as independent observations. For each dataset with a unique highest-performing configuration, this configuration was compared with the configuration achieving the second-highest accuracy on the same 200 test instances. If multiple configurations shared the second-highest accuracy, each was compared separately with the highest-performing configuration. No significance test was conducted when multiple configurations shared the highest accuracy, since in these cases no unique highest-performing configuration existed for the predefined best-versus-second-best comparison. The exact McNemar test was used for all pairwise comparisons as some comparisons contained only a small number of instances for which the two configurations differed in correctness. To account for multiple comparisons, the resulting $p$-values were adjusted using the Holm procedure \cite{holm1979}. Statistical significance was assessed at the conventional threshold of $\alpha=0.05$.

\section{Results}
\label{sec:results}
    
Table~\ref{tab:results_combined} summarizes the accuracy scores for all prompting techniques across the benchmark datasets and evaluated configurations. Zero-Shot Prompting is reported as a single configuration as it does not use additional task-specific examples. For Few-Shot Prompting, 100, 200, and 300 refer to the number of demonstration examples, while for PE2 they refer to the number of examples used for optimization. For RAG, these numbers refer to the number of examples available for retrieval, of which five were retrieved for each evaluation instance.
    
Few-Shot Prompting achieved its highest average accuracy of 77.83\% with 200 demonstration examples. RAG showed the clearest positive scaling behavior as the number of examples available for retrieval increased, reaching the highest overall average accuracy of 78.33\% with 300 examples.
    
The highest individual result across all experiments was achieved by Zero-Shot Prompting on BoolQ with 90.0\% accuracy. In contrast, the best results for arithmetic and logical reasoning tasks were generally obtained with Few-Shot Prompting or RAG using larger numbers of examples. RAG with 300 examples available for retrieval achieved the highest observed accuracy on both Elementary Math QA (66.0\%) and Logical Fallacy Detection (85.5\%).
    
The results further show that no prompting technique consistently outperformed all others across all reasoning domains. Few-Shot Prompting achieved comparatively high accuracies on structured reasoning tasks such as Navigate and Logical Fallacy Detection. On Fact Checker, Few-Shot Prompting with 200 demonstration examples and RAG with 200 or 300 examples available for retrieval jointly achieved the highest accuracy of 86.5\%. PE2 achieved the best performance on Social IQA with 79.0\% accuracy but did not outperform the other prompting techniques on average.
    
Increasing the number of task-specific examples affected the prompting techniques differently. Few-Shot Prompting improved slightly from 76.92\% to 77.83\% when the number of demonstration examples increased from 100 to 200 but did not improve further with 300 examples. In contrast, RAG showed a more consistent scaling trend, improving from 75.75\% to 78.33\% as the number of examples available for retrieval increased from 100 to 300. PE2 remained comparatively stable as the number of examples used for optimization increased and did not show a clear positive scaling effect.
    
Overall, the results show substantial variation in the relative performance of the prompting techniques across datasets. Few-Shot Prompting and RAG achieved the highest average accuracies in several configurations, while Zero-Shot Prompting remained competitive on some datasets and PE2 achieved the highest accuracy on Social IQA.

The statistical comparisons showed that none of the observed differences between the highest-performing and second-highest-performing configurations in Table~\ref{tab:results_combined} were statistically significant. Across the eight pairwise comparisons, the smallest unadjusted $p$-value was obtained for Navigate, where Few-Shot Prompting with 300 examples (71.5\% accuracy) was compared with Few-Shot Prompting using 100 and 200 examples (69.5\% accuracy in both cases; exact McNemar $p=0.219$ in both comparisons). The other six comparisons yielded $p=1.000$. After correcting the eight comparisons for multiple testing using the Holm procedure, all adjusted $p$-values were 1.000. Thus, the observed numerical differences in accuracy between the compared configurations were not statistically significant at $\alpha=0.05$.

\section{Discussion}
   \label{sec:discussion}
    
The results demonstrate that no prompting technique consistently outperformed all others across all reasoning domains. Instead, the effectiveness of Zero-Shot Prompting, Few-Shot Prompting, RAG, and PE2 strongly depended on the structure and characteristics of the benchmark tasks. None of the differences between the highest-performing and second-highest-performing configurations reached statistical significance. The observed accuracy differences should therefore be interpreted as descriptive differences within the present evaluation rather than as evidence that one configuration is statistically superior to another. The observed task dependence is consistent with prior work showing that prompting performance varies substantially across tasks, prompt formulations, and reasoning requirements \cite{brown2020languagemodels,srivastava2023beyond,schulhoff2025promptreport}.
    
Few-Shot Prompting achieved its highest average accuracy of 77.83\% with 200 demonstration examples and comparatively high accuracies on Logical Fallacy Detection and Navigate. On these tasks, fixed demonstrations may help communicate task-specific input-output patterns and expected answer formats. This interpretation is consistent with prior work showing that in-context examples can substantially affect LLM task performance \cite{brown2020languagemodels,liu2022knn}.
    
However, the scaling behavior of Few-Shot Prompting was inconsistent. Increasing the number of demonstration examples from 100 to 200 slightly improved average performance, while further increasing the number to 300 did not produce additional gains. This suggests that the effectiveness of Few-Shot Prompting depends not only on the number of examples but also on their relevance and compatibility with the evaluation task. Previous studies similarly showed that example selection and prompt composition can significantly influence Few-Shot performance \cite{liu2022knn,su2022select}.
    
RAG showed the clearest positive trend in average accuracy as the number of examples available for retrieval increased, reaching the highest overall average accuracy with a pool of 300 examples. The largest gains from the 100- to 300-example retrieval pools occurred on Logical Fallacy Detection and Elementary Math QA, where accuracy increased by 6.5 and 5.5 absolute percentage points, respectively. This pattern is consistent with the idea that a larger retrieval pool can increase the opportunity to retrieve relevant examples, although the effect was task-dependent. On BoolQ, accuracy decreased as the retrieval pool increased, while on Navigate RAG remained below the best Few-Shot result. Thus, retrieval-based example augmentation did not provide uniform gains across tasks. This task dependence is consistent with prior work emphasizing the importance of the quality and relevance of retrieved context \cite{lewis2020rag,borgeaud2022retro}.
    
Zero-Shot Prompting achieved the highest result on BoolQ with 90.0\% accuracy. BoolQ provides the passage needed to answer each binary question within the input itself \cite{clark2019boolq}. In the present experiments, neither fixed demonstration examples nor dynamically retrieved examples improved on the Zero-Shot result. This suggests that additional contextual examples are not necessarily beneficial when the information required for the decision is already contained in the evaluation instance.
    
PE2 produced comparatively stable average results across the three optimization sizes, with average accuracies of 74.08\%, 74.25\%, and 74.00\% for 100, 200, and 300 optimization examples, respectively. Although it did not exceed the best Few-Shot or RAG configurations on average, PE2 achieved the highest accuracy on Social IQA. Increasing the number of examples used for optimization therefore did not translate into higher average performance in the present setup. The PE2-inspired optimization was limited to three iterations, so these findings characterize automated prompt optimization under the specific search procedure and optimization budget evaluated in this study rather than the potential of automated prompt optimization more generally.
    
Overall, the findings suggest that prompting techniques should not be treated as universally interchangeable optimization strategies. Instead, their relative effectiveness varies across the evaluated tasks and depends on how contextual examples are incorporated. Few-Shot Prompting achieves comparatively high accuracies on several structured reasoning tasks, while the evaluated RAG configuration shows higher average accuracy with a larger pool of examples available for retrieval. Zero-Shot Prompting remains competitive and achieves the highest result on BoolQ, whereas the PE2-inspired approach does not outperform Few-Shot Prompting or RAG on average. These differences reinforce the need to evaluate prompting techniques in relation to the characteristics of the target task rather than assuming that a more complex prompting strategy will consistently improve performance.
    
    \subsection{Limitations}
    
Several limitations should be considered when interpreting the results of this study. First, GPT-4o-mini was used as a representative LLM and held constant across all experiments to enable a controlled comparison of prompting techniques. Consequently, the findings characterize prompting behavior for this LLM and may not generalize directly to other LLM architectures or model sizes.
    
Second, the evaluation was limited to six benchmark datasets and 200 test instances per dataset. While this setup enabled controlled comparisons across prompting techniques, larger evaluation sets and additional domains could produce more robust conclusions regarding prompting behavior across tasks. In addition, each configuration was evaluated once per test instance without repeated generations. Although a temperature of 0.0 was used consistently across all experiments, repeated runs could provide additional insight into potential output variability. The statistical comparisons therefore assess differences in paired correctness outcomes on the evaluated test instances rather than variability across repeated LLM generations.
    
    Third, the evaluation used accuracy as the common performance metric across all six datasets to enable consistent cross-domain comparisons. However, the test sets were not necessarily perfectly balanced across answer classes, which can affect the interpretation of accuracy. Moreover, accuracy does not capture class-specific error patterns or qualitative differences in reasoning behavior.
    
In addition, the implemented RAG setup retrieved task-specific examples rather than information from an external knowledge base. The experiments therefore evaluated retrieval-based example augmentation rather than knowledge-grounded RAG.
    
Finally, the PE2 implementation was adapted specifically for the experimental setup of this study. Differences in optimization strategy, search space design, or meta-prompt configuration may influence the observed results and could lead to different outcomes in alternative implementations.
    
\subsection{Implications}
    
The findings have practical implications for the design of LLM-based systems. Rather than selecting prompting techniques based solely on their complexity, practitioners should evaluate them on the target task. The results also show that providing more task-specific examples does not necessarily improve performance: Few-Shot Prompting achieved its highest average accuracy with 200 demonstration examples, whereas RAG continued to improve on average as the pool of examples available for retrieval increased to 300. Consequently, both the choice of prompting technique and the amount of contextual information should be treated as task-dependent design decisions rather than fixed defaults.

\section{Conclusion and Future Work}
\label{sec:conclusion}

This section summarizes the main findings and outlines directions for future research.
    
\subsection{Conclusion}
    
Our paper compared Zero-Shot Prompting, Few-Shot Prompting, Retrieval-Augmented Generation (RAG), and PE2-inspired automatic prompt optimization across multiple reasoning domains under a controlled experimental setup. All prompting techniques were evaluated using the same LLM, benchmark datasets, evaluation metric, and inference settings in order to ensure direct comparability.
    
The results showed that no prompting technique consistently outperformed all others across all benchmark datasets. None of the pairwise differences between the highest-performing and second-highest-performing configurations reached statistical significance after Holm correction for multiple comparisons. Few-Shot Prompting achieved comparatively high accuracies on several structured reasoning tasks, while RAG showed the clearest positive trend in average accuracy as the number of examples available for retrieval increased. Zero-Shot Prompting remained competitive and achieved the highest accuracy on BoolQ. The PE2-inspired approach achieved the highest accuracy on Social IQA but did not outperform the best Few-Shot or RAG configurations on average.
    
Overall, the findings indicate that prompting effectiveness strongly depends on task structure and reasoning domain. Prompting techniques should therefore not be treated as universally optimal strategies but instead selected according to the characteristics and contextual requirements of the target task.
    
    \subsection{Future Work}
    
    Future research should extend the evaluation to additional reasoning domains, benchmark datasets, and LLM architectures in order to improve the generalizability of the findings. In particular, evaluating prompting techniques on larger and more diverse datasets may provide deeper insights into domain-specific prompting behavior.
    
    Furthermore, future work could investigate more advanced prompting strategies such as Chain-of-Thought Prompting, Few-Shot Chain-of-Thought, or hybrid approaches combining retrieval mechanisms with automated prompt optimization. Exploring adaptive retrieval strategies, example selection methods, and larger prompt optimization search spaces may further improve prompting robustness and performance across complex reasoning tasks.
    
    Finally, future studies could complement quantitative evaluation metrics with qualitative analyses of reasoning behavior, explanation consistency, and robustness under noisy or ambiguous inputs in order to better understand the strengths and limitations of different prompting approaches.
    
    \bibliographystyle{IEEEtran}
    \bibliography{references.bib}

@inproceedings{brown2020languagemodels,
  title={{Language Models are Few-Shot Learners}},
  author={Brown, Tom B. and Mann, Benjamin and Ryder, Nick and Subbiah, Melanie and Kaplan, Jared and Dhariwal, Prafulla and Neelakantan, Arvind and Shyam, Pranav and Sastry, Girish and Askell, Amanda and Agarwal, Sandhini and Herbert-Voss, Ariel and Krueger, Gretchen and Henighan, Tom and Child, Rewon and Ramesh, Aditya and Ziegler, Daniel M. and Wu, Jeffrey and Winter, Clemens and Hesse, Christopher and Chen, Mark and Sigler, Eric and Litwin, Mateusz and Gray, Scott and Chess, Benjamin and Clark, Jack and Berner, Christopher and McCandlish, Sam and Radford, Alec and Sutskever, Ilya and Amodei, Dario},
  booktitle={{The 34th Conference on Neural Information Processing Systems}},
  pages={1877--1901},
  year={2020}
}

@inproceedings{lewis2020rag,
  title={{Retrieval-Augmented Generation for Knowledge-Intensive NLP Tasks}},
  author={Lewis, Patrick and Perez, Ethan and Piktus, Aleksandra and Petroni, Fabio and Karpukhin, Vladimir and Goyal, Naman and K{\"u}ttler, Heinrich and Lewis, Mike and Yih, Wen-tau and Rockt{\"a}schel, Tim and Riedel, Sebastian and Kiela, Douwe},
  booktitle={{The 34th Conference on Neural Information Processing Systems}},
  pages={9459--9474},
  year={2020}
}

@article{liu2023pretrain,
  title={{Pre-train, Prompt, and Predict: A Systematic Survey of Prompting Methods in Natural Language Processing}},
  author={Liu, Pengfei and Yuan, Weizhe and Fu, Jinlan and Jiang, Zhengbao and Hayashi, Hiroaki and Neubig, Graham},
  journal={{ACM Computing Surveys}},
  volume={55},
  number={9},
  pages={1--35},
  year={2023},
  publisher={{Association for Computing Machinery}},
  doi={10.1145/3560815}
}

@article{chang2024survey,
  title={{A Survey on Evaluation of Large Language Models}},
  author={Chang, Yupeng and Wang, Xu and Wang, Jindong and Wu, Yuan and Yang, Longyue and Zhu, Kaijie and Chen, Hao and Yi, Xiaoyuan and Wang, Cunxiang and Wang, Yidong and Ye, Wei and Zhang, Yue and Chang, Yi and Yu, Philip S. and Yang, Qiang and Xie, Xing},
  journal={{ACM Transactions on Intelligent Systems and Technology}},
  volume={15},
  number={3},
  pages={1--45},
  year={2024},
  publisher={{Association for Computing Machinery}},
  doi={10.1145/3641289}
}

@article{schulhoff2025promptreport,
  title={{The Prompt Report: A Systematic Survey of Prompt Engineering Techniques}},
  author={Schulhoff, Sander and Ilie, Michael and Balepur, Nishant and Kahadze, Konstantine and Liu, Amanda and Si, Chenglei and Li, Yinheng and Gupta, Aayush and Han, Han and Schulhoff, Shreya and Dulepet, Pranav S. and Vidyadhara, Saurav and Ki, David and Agrawal, Sweta and Pham, Chau and Kroiz, Gera and Li, Fei and Tao, Haoran and Srivastava, Aarnav and Resnik, Philip},
  journal={{arXiv preprint arXiv:2406.06608}},
  year={2025}
}

@inproceedings{ye2024promptengineer,
  title={{Prompt Engineering a Prompt Engineer}},
  author={Ye, Qinyuan and Ahmed, Mohamed and Pryzant, Reid and Khani, Fereshte},
  booktitle={{Findings of the Association for Computational Linguistics: ACL 2024}},
  pages={355--385},
  year={2024},
  publisher={{Association for Computational Linguistics}},
  doi={10.18653/v1/2024.findings-acl.21}
}

@article{openai2023gpt4,
  title={{GPT-4 Technical Report}},
  author={{OpenAI}},
  journal={{arXiv preprint arXiv:2303.08774}},
  year={2023},
  doi={10.48550/arXiv.2303.08774}
}

@inproceedings{wei2022chain,
  title={{Chain-of-Thought Prompting Elicits Reasoning in Large Language Models}},
  author={Wei, Jason and Wang, Xuezhi and Schuurmans, Dale and Bosma, Maarten and Ichter, Brian and Xia, Fei and Chi, Ed H. and Le, Quoc V. and Zhou, Denny},
  booktitle={{The 36th Conference on Neural Information Processing Systems}},
  year={2022}
}

@article{su2022select,
  title={{Selective Annotation Makes Language Models Better Few-Shot Learners}},
  author={Su, Hongjin and Kasai, Jungo and Wu, Chen Henry and Shi, Weijia and Wang, Tianlu and Xin, Jiayuan and Zhang, Rui and Ostendorf, Mari and Zettlemoyer, Luke and Smith, Noah A. and Yu, Tao},
  journal={arXiv preprint arXiv:2209.01975},
  year={2022}
}

@inproceedings{liu2022knn,
title={{What Makes Good In-Context Examples for GPT-3?}},
  author={Liu, Jiachang and Shen, Dinghan and Zhang, Yizhe and Dolan, Bill and Carin, Lawrence and Chen, Weizhu},
  booktitle={The 3rd Workshop on Deep Learning Inside Out: Knowledge Extraction and Integration for Deep Learning Architectures},
  pages={100--114},
  year={2022},
  publisher={Association for Computational Linguistics},
  doi={10.18653/v1/2022.deelio-1.10}
}

@inproceedings{zhou2022humanlevel,
  title={{Large Language Models are Human-Level Prompt Engineers}},
  author={Zhou, Yongchao and Muresanu, Andrei Ioan and Han, Ziwen and Paster, Keiran and Pitis, Silviu and Chan, Harris and Ba, Jimmy},
  booktitle={{The Eleventh International Conference on Learning Representations}},
  year={2023}
}

@inproceedings{pryzant2023automatic,
  title={{Automatic Prompt Optimization with ``Gradient Descent'' and Beam Search}},
  author={Pryzant, Reid and Iter, Dan and Li, Jerry and Lee, Yin Tat and Zhu, Chenguang and Zeng, Michael},
  booktitle={{The 2023 Conference on Empirical Methods in Natural Language Processing}},
  pages={7957--7968},
  year={2023},
  publisher={{Association for Computational Linguistics}},
  doi={10.18653/v1/2023.emnlp-main.494}
}

@inproceedings{borgeaud2022retro,
  title={{Improving Language Models by Retrieving from Trillions of Tokens}},
  author={Borgeaud, Sebastian and Mensch, Arthur and Hoffmann, Jordan and Cai, Trevor and Rutherford, Eliza and Millican, Katie and van den Driessche, George and Lespiau, Jean-Baptiste and Damoc, Bogdan and Clark, Aidan and de Las Casas, Diego and Guy, Aurelia and Menick, Jacob and Ring, Roman and Hennigan, Tom and Huang, Saffron and Maggiore, Loren and Jones, Chris and Cassirer, Albin and Sifre, Laurent},
  booktitle={{The 39th International Conference on Machine Learning}},
  pages={2206--2240},
  year={2022},
  publisher={{PMLR}}
}

@article{srivastava2023beyond,
  title={Beyond the Imitation Game: Quantifying and Extrapolating the Capabilities of Language Models},
  author={Srivastava, Aarohi and Rastogi, Abhinav and Rao, Abhishek and others},
  journal={Transactions on Machine Learning Research},
  year={2023}
}

@inproceedings{clark2019boolq,
  title={{BoolQ: Exploring the Surprising Difficulty of Natural Yes/No Questions}},
  author={Clark, Christopher and Lee, Kenton and Chang, Ming-Wei and Kwiatkowski, Tom and Collins, Michael and Toutanova, Kristina},
  booktitle={{The 2019 Conference of the North American Chapter of the Association for Computational Linguistics: Human Language Technologies}},
  pages={2924--2936},
  year={2019},
  publisher={{Association for Computational Linguistics}},
  doi={10.18653/v1/N19-1300}
}

@inproceedings{sap2019socialiqa,
  title={{SocialIQA: Commonsense Reasoning about Social Interactions}},
  author={Sap, Maarten and Rashkin, Hannah and Chen, Derek and LeBras, Ronan and Choi, Yejin},
  booktitle={{The 2019 Conference on Empirical Methods in Natural Language Processing and the 9th International Joint Conference on Natural Language Processing}},
  pages={4463--4473},
  year={2019},
  publisher={{Association for Computational Linguistics}},
  doi={10.18653/v1/D19-1454}
}

@misc{openai2024gpt4o,
  title={{Hello GPT-4o}},
  author={{OpenAI}},
  year={2024},
  howpublished={\url{https://openai.com/index/hello-gpt-4o/}},
  note={Accessed: 2026-05-15}
}

@inproceedings{lhoest2021datasets,
  title={{Datasets: A Community Library for Natural Language Processing}},
  author={Lhoest, Quentin and Delangue, Cl{\'e}ment and Matussi{\`e}re, Guillaume and von Platen, Patrick and Drame, Mariama and Jernite, Yacine and Bekman, Stas and Cistac, Pierric and Goehringer, Julien and Mustar, Victor and others},
  booktitle={The 2021 Conference on Empirical Methods in Natural Language Processing: System Demonstrations},
  pages={175--184},
  year={2021}
}

@INPROCEEDINGS{fengEtAlRAGs,
  author={Feng, Junzhe and Ariannejad, MohammadMahdi and Idrus, Intan Izafina and Goh, Zhe Kai and Tan, Jian Ding and Kang, Chia Chao and Byuiyan, Mohammad Arif Sobhan and Tee, Wei Hown},
  booktitle={The 9th International Conference on Recent Advances and Innovations in Engineering (ICRAIE)}, 
  title={{Retrieval-Augmented Generation: A Comprehensive Survey of Foundations, Evolution, Applications, Limitations, and Future Directions}}, 
  year={2025},
  volume={},
  number={},
  pages={148-153}
}

@article{mcnemar1947,
  author  = {McNemar, Quinn},
  title   = {{Note on the Sampling Error of the Difference Between Correlated Proportions or Percentages}},
  journal = {Psychometrika},
  volume  = {12},
  number  = {2},
  pages   = {153--157},
  year    = {1947},
  doi     = {10.1007/BF02295996}
}

@article{westfall2010multiple,
  author  = {Westfall, Peter H. and Troendle, James F. and Pennello, Gene},
  title   = {{Multiple McNemar Tests}},
  journal = {Biometrics},
  year    = {2010},
  doi     = {10.1111/j.1541-0420.2010.01408.x}
}

@article{holm1979,
  author  = {Holm, Sture},
  title   = {{A Simple Sequentially Rejective Multiple Test Procedure}},
  journal = {Scandinavian Journal of Statistics},
  volume  = {6},
  number  = {2},
  pages   = {65--70},
  year    = {1979}
}

@inproceedings{Schlippe2026RAGStrategies6i2k,
  author = {Tim Schlippe and Simon Martin and Michael Marketsmüller},
  title = {{RAG Strategies for Natural Language-Based SQL Query and REST API Call Generation}},
  booktitle = {The 3rd International Conference on Foundation and Large Language Models (FLLM)},
  address = {Barcelona, Spain}, 
  year = {2026}
}

@inproceedings{Kivroglou2025InvestigatingRetrieval22wm,
  author = {Paraskevi Kivroglou and Tim Schlippe and Simon Martin},
  title = {{Investigating Retrieval Augmented Generation for LLM-Based Code Generation}},
  booktitle = {The 3rd International Conference on Foundation and Large Language Models (FLLM)},
  address = {Vienna, Austria},
  year = {2025},
  url = {https://ieeexplore.ieee.org/document/11391177}
}

@inproceedings{tejaswi2025rare,
  title={{RARe: Retrieval Augmented Retrieval with In-Context Examples}},
  author={Tejaswi, Atula and Lee, Yoonsang and Sanghavi, Sujay and Choi, Eunsol},
  booktitle={Conference on Language Modeling},
  year={2025}
}
    
    \end{document}